\documentclass[lettersize,journal]{IEEEtran}
\usepackage{amsmath,amsfonts}
\usepackage{algorithmic}
\usepackage{algorithm}
\usepackage{array}
\usepackage[caption=false,font=normalsize,labelfont=sf,textfont=sf]{subfig}
\usepackage{textcomp}
\usepackage{stfloats}
\usepackage{url}
\usepackage{verbatim}
\usepackage{graphicx}
\usepackage{cite}
\usepackage{booktabs}
\usepackage{balance}

\usepackage[colorlinks=true, linkcolor=black, citecolor=blue, urlcolor=blue]{hyperref}
\begin{document}

\title{Physics-Aware 3D Gaussian Editing for Driving Scene Generation}
\author{
Feng Zhou$^{1,3}$,
Jian Zhang$^{2,3}$,
Yuhang Sun$^{3}$,
He Wang$^{1}$,
Qiong Wen$^{3}$,
Debao Kong$^{3}$,
Tieru Wu$^{1,*}$,
Rui Ma$^{1,*}$%
\thanks{$^{1}$School of Artificial Intelligence, Jilin University, Changchun, China.}
\thanks{$^{2}$National Key Laboratory of Automotive Chassis Integration and Bionics, Jilin University, Changchun, China.}
\thanks{$^{3}$China FAW Group Co., Ltd., Changchun, China.}
\thanks{$^{*}$Corresponding authors}
}

\maketitle

\begin{abstract}
3D Gaussian Splatting (3DGS) has shown great potential in autonomous driving simulation and data generation, enabling photorealistic reconstruction and flexible scene manipulation. However, existing 3DGS scene editing methods have limited support for road geometry editing (e.g., inserting speed humps or sunken roads), and generally do not couple such edits with plausible vehicle-road interaction dynamics. Such editing is essential for generating training data under extreme driving scenarios or evaluating system reliability under these road irregularities. Moreover, many optimization-based methods require minutes of per-edit refinement, while existing efficient alternatives mainly focus on appearance-level or object-level manipulation rather than physics-aware road irregularity editing. To address these limitations, we propose RoVES, a Road-and-Vehicle Editing System for physics-aware 3D Gaussian editing in driving scenes. RoVES enables single-image-driven road geometry insertion and couples the edited road profile with a 4-DOF half-car vehicle dynamics model to achieve physics-aware vehicle pose correction in vertical displacement and pitch. RoVES inserts road elements in a one-shot, optimization-free pipeline (1.84\,s), and the full pipeline (including color transfer and vehicle-dynamics-based pose correction) completes in 6.24\,s; it edits dynamic vehicles via pose editing and corrects poses frame-by-frame to approximate dynamics-consistent vertical displacement and pitch responses. Experiments on the Waymo dataset show that RoVES provides practical efficiency and competitive visual consistency for physics-aware driving scene generation.
\end{abstract}


\section{Introduction}
\IEEEPARstart{T}{he} development of autonomous driving technology relies heavily on massive training and testing data under diverse and challenging conditions to train robust models and verify safety assurance \cite{li2024choose}. Among these conditions, discrete road structures such as speed humps, potholes and sunken roads are critical long-tail scenarios: they induce vertical displacement and pitch of the vehicle body, causing pose jitter and sensor data offsets that degrade perception, localization and control, and may even lead to instability accidents \cite{sadeghi2024investigating}. However, real‑world collection of such data is costly, risky and sparse, making real‑scene 3D reconstruction with controllable editing a scalable path to generate both training data and simulation environments for these critical conditions.

3DGS\cite{kerbl20233d} has become a widely adopted representation for driving scene reconstruction, enabling photorealistic reconstructions of static and dynamic elements \cite{yan2024street, liu2025omni}. However, existing 3DGS editing methods have not fully addressed two critical aspects required by physics-aware driving scene generation. First, current driving-scene editing methods mainly focus on appearance modification, weather simulation, trajectory editing, or object-level manipulation, but they do not explicitly model metric road-height irregularities as excitation for vehicle-road interaction dynamics \cite{xiong2025drivinggaussian++, wang2026horizonforge, liu2026horizonweaver}. Consequently, edited road irregularities cannot induce dynamics-consistent vehicle responses, limiting their suitability for physics-aware driving scene generation. Second, iterative gradient-based methods often require minutes of per-edit optimization \cite{haque2023instruct, chen2024gaussianeditor, wu2024gaussctrl}, while efficient direct-editing alternatives \cite{ye2024gaussian, jain2025gaussiancut} mainly support object-level operations such as translation, rotation, deletion, or insertion, and lack the specialized functionality needed for road-geometry insertion with vehicle-pose correction. Although recent work has integrated 3DGS with physics for rigid-object manipulation in indoor scenes (e.g., RoboSimGS~\cite{zhao2026high} which adopts a hybrid representation to handle rigid and articulated objects), such coupling for driving scenes—specifically continuous road geometry editing and vehicle vertical dynamics—remains underexplored because the latter imposes fundamentally different requirements on geometric representation and vehicle-response modeling.


\begin{figure*}[!tb]
\vspace*{-1.0em}
\centering
\includegraphics[width=1.0 \linewidth]{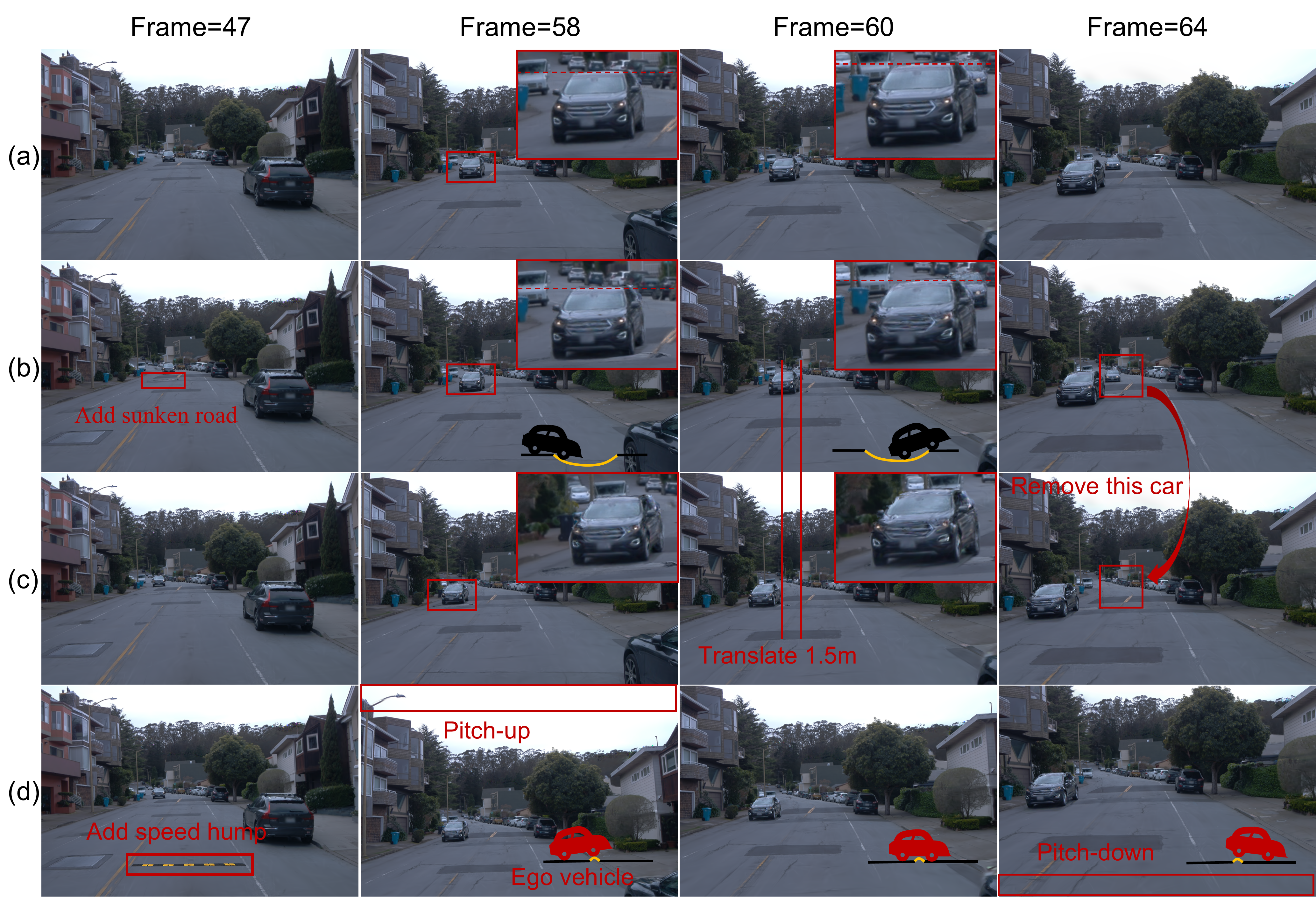} 
\caption{Editing results of RoVES. (a) Original view; (b)--(d) demonstrate progressively more complex editing capabilities achieved by our method: (b) Road editing with physics-aware vehicle response; (c) Dynamic vehicle pose editing; (d) Ego-view road editing with physics-aware ego-vehicle pose correction.}
\label{fig:overview}
\end{figure*}

To fill these gaps, we propose RoVES, a novel 3DGS road and vehicle editing framework that explicitly integrates vehicle-road interaction dynamics for physics-aware scene editing. RoVES adopts an optimization-free editing architecture that couples a half-car dynamics model with edited road geometry to perform physics-aware, dynamics-consistent vehicle pose correction, while inserting static road elements in a single forward pass and manipulating dynamic vehicles via pose editing. Example editing results are shown in Fig.~\ref{fig:overview}. The core contributions of this paper are as follows:
\begin{itemize}
\item A physics-aware coupling between 3DGS road editing and a classic vehicle dynamics model: the edited road height profile provides physical excitation, and frame-wise pose correction yields dynamics-consistent vertical displacement and pitch responses, linking visual editing with physics-aware pose correction.
\item An optimization-free, geometry-pose decoupled editing framework that separates static road Gaussian insertion from dynamic vehicle pose adjustment, requiring no backpropagation or fine-tuning (1.84\,s object insertion, 6.24\,s full pipeline).
\item A single-image and text-driven insertion pipeline with KNN adaptive scale and Lab color transfer that suppresses monocular point-cloud artifacts for high-fidelity fusion.
\end{itemize}

Extensive comparative and ablation studies on the Waymo dataset indicate that our method provides practical efficiency and physics-aware vehicle responses, with competitive geometric and visual quality.

\begin{figure*}[!tb]
\vspace*{-1.0em}
\centering
\includegraphics[width=1.0 \textwidth]{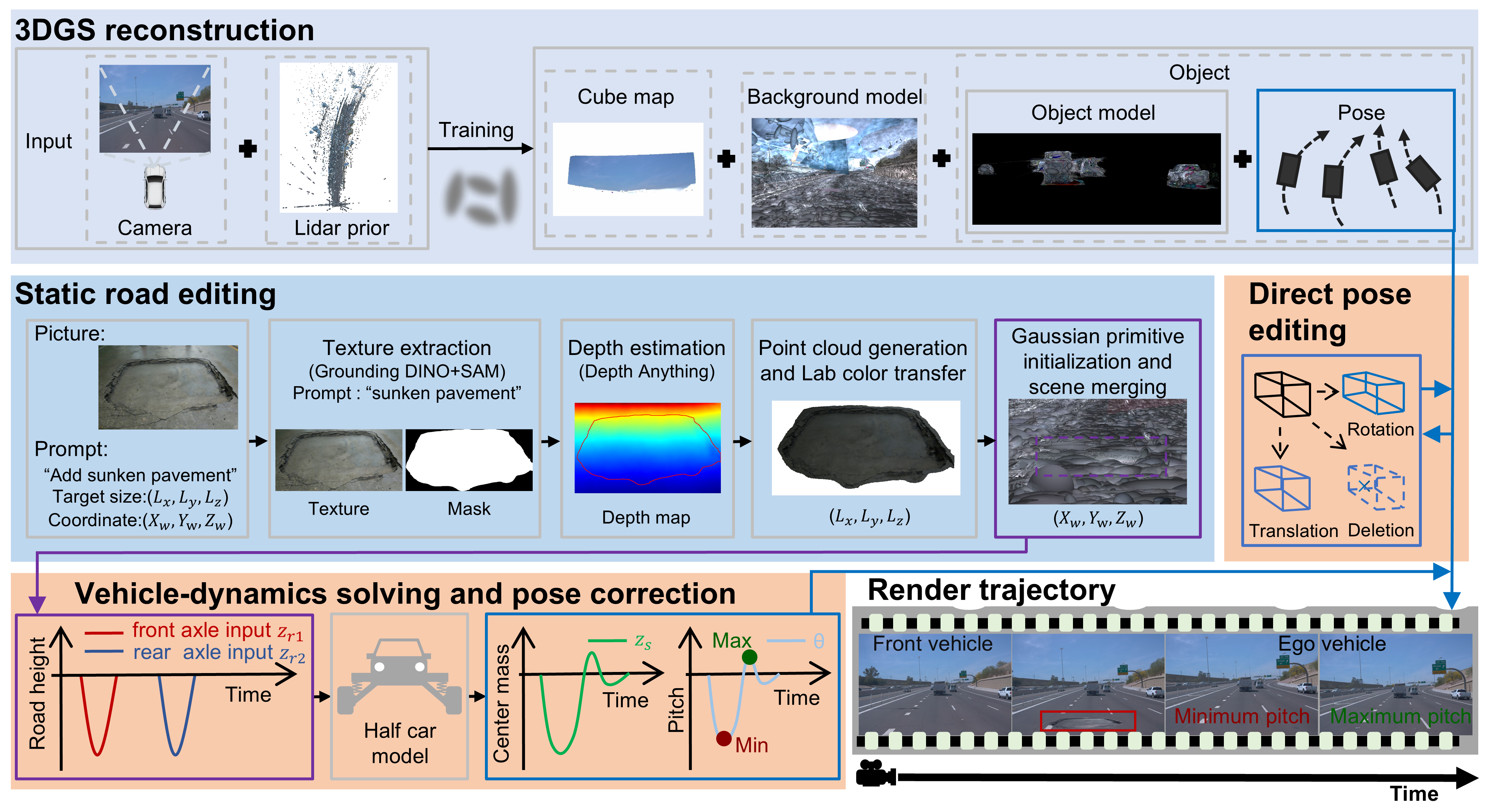}
\caption{Overview of our pipeline. The scene is reconstructed into static background, dynamic objects, and sky. Our framework decouples into two branches: (static) road geometry elements inserted from a single image and text prompt using detection-segmentation, monocular depth lifting, and scale-corrected Gaussian merging in a one-shot forward pass; (dynamic) vehicles support direct pose editing (translation, rotation, deletion) and physics-aware pose correction via vehicle dynamics: a half-car dynamics model uses the edited road profile to compute dynamics-consistent vertical displacement and pitch responses. Ego-centric videos are then rendered.}
\label{fig:pipeline}
\end{figure*}

\section{Related Work}

We review related work from two directions: 3DGS scene editing and vehicle-road interaction physics for data generation.
\subsection{3D Gaussian Splatting Scene Editing Methods}
3DGS scene editing methods are broadly categorized into optimization-based and optimization-free approaches, representing a fundamental trade-off between flexibility and efficiency.

Optimization-based methods, the current mainstream, iteratively fine-tune Gaussian parameters via rendering loss. Full-scene approaches such as GaussianEditor \cite{chen2024gaussianeditor}, GaussCtrl \cite{wu2024gaussctrl} and 3DSceneEditor \cite{yan20243dsceneeditor} offer high flexibility through text- or image-driven loss guidance, but require 2--10 minutes per edit due to the costly gradient-based optimization. Local methods like On-the-Fly GS \cite{xu2025gaussian} and EditSplat \cite{lee2025editsplat} reduce runtime to 1--3 minutes by confining updates to user-specified regions, yet the iterative refinement still cannot match the batch generation throughput demanded by autonomous driving simulation.

Optimization-free methods achieve second-level editing by eliminating gradient optimization and relying solely on object-level Gaussian decoupling and geometric transformations. Gaussian Grouping \cite{ye2024gaussian} and GaussianCut \cite{jain2025gaussiancut} enable instant translation, rotation, and deletion without retraining, though they depend on high-quality instance segmentation (which degrades in occluded driving scenes) and only support basic pose adjustments and object addition/deletion.

\subsection{Vehicle-Road Interaction Physics and Data Generation}
Autonomous driving simulation benefits from visually realistic scenes and plausible vehicle-road interaction dynamics for algorithm verification and data generation. Mainstream platforms such as CarSim, Prescan, and CARLA provide mature multi-degree-of-freedom vehicle dynamics \cite{pahk2024lane, lin2022model}, but they suffer from a significant Sim-to-Real gap due to manually modeled scenes that lack real-world texture and geometric fidelity, limiting their usefulness for generating realistic training data. 

Real-road reconstruction-based simulation (e.g., using NeRF/3DGS) improves visual realism, yet most works focus solely on rendering and trajectory planning, neglecting dynamics induced by road roughness \cite{feng2025gaussian, ma2026fastphysgs}. Existing kinematics simulations (e.g., lane changing, translation) on reconstructed scenes also ignore vertical displacement and sensor jitter \cite{tian2025simscale, gao2025rad}. Recent efforts like RoboSimGS \cite{zhao2026high} integrate 3DGS with physics for robotic manipulation, but their hybrid representation (3DGS background + mesh objects) is inefficient for large-scale outdoor road editing and does not address vehicle dynamics, thus cannot generate road-editing data with physics-aware, dynamics-consistent vehicle responses.

In summary, existing methods still lack an efficient framework that jointly supports realistic road editing, rapid batch generation, and physics-aware vehicle response modeling—the gap addressed by RoVES.

\section{Methodology}
Our editing framework explicitly integrates vehicle-road interaction dynamics for physics-aware scene editing. Specifically, a half-car dynamics model is coupled with the edited road geometry: the road height profile serves as the external excitation input to the model, and the solved body displacement and pitch are superimposed onto vehicle poses. To realize this coupling efficiently, we build RoVES around a \textbf{geometry--pose decoupled} framework (Fig.~\ref{fig:pipeline}) that enables optimization-free editing: static road structures are inserted through a one-shot pipeline; dynamic vehicles are manipulated in two modes---direct pose editing (translation, rotation, deletion of object instances) by simply updating pose sequences, and physics-aware pose correction via vehicle dynamics as described above. The entire process requires no retraining.

\subsection{3DGS Scene Representation}
\label{sec:recon}

We reconstruct driving scenes from the Waymo Open Dataset \cite{sun2020scalability} using the Street Gaussians \cite{yan2024street} framework, which decouples the scene into static background, dynamic objects, and sky branches. After one-time reconstruction, all Gaussian parameters (position, covariance, color, opacity) are frozen; only per-frame object poses are retained independently. This frozen, pose-decoupled representation is the prerequisite for our editing framework, described next.

\subsection{Geometry--Pose Decoupled Editing Framework}
\label{sec:paradigm}
Since the core editing tasks in driving scenes mainly involve spatial pose changes, we adopt an optimization-free strategy. As illustrated in Fig.~\ref{fig:comparison}, traditional pipelines (left) iteratively optimize Gaussian parameters for every edit, costing minutes. Our framework (right) decouples editing into two independent branches that operate directly on the frozen pre-trained scene:

\begin{figure}[!tb]
\centering
\includegraphics[width=1\linewidth]{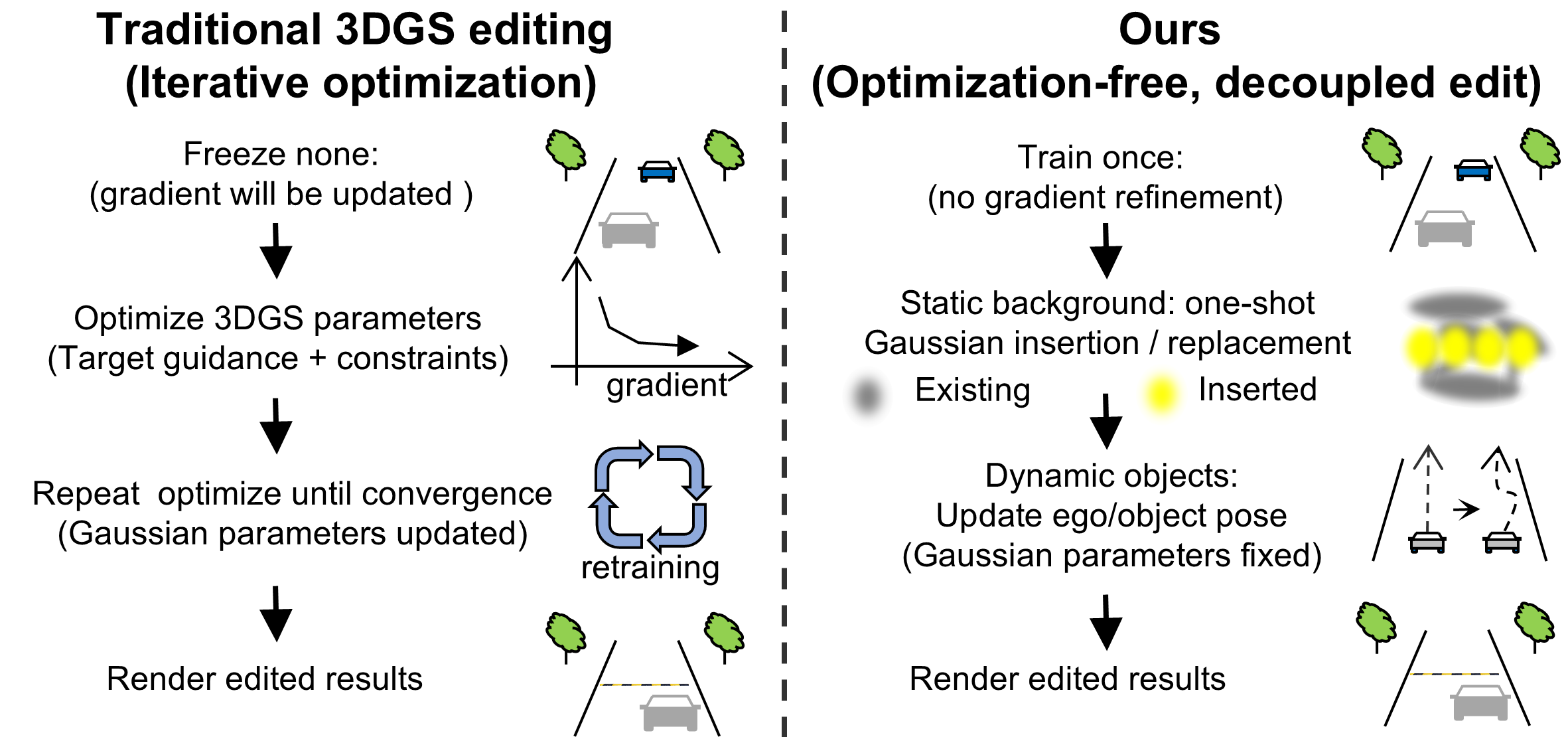}
\caption{Comparison of editing pipelines. Left: traditional iterative optimization converges per edit. Right: our optimization-free approach decouples static one-shot insertion from dynamic pose updates, requiring no retraining.}
\label{fig:comparison}
\end{figure}

\textit{Static structure editing.} For road structures not modeled as independent dynamic entities, a one-shot forward pass performs geometric generation and Gaussian primitive initialization on the ROI region of the input image. The resulting Gaussians are merged into the background by replacing primitives within the projected area, without any gradient backpropagation. See Section~\ref{sec:static} for details on depth estimation, mask projection, and appearance alignment.

\textit{Dynamic object editing.} Dynamic editing covers two cases: pose editing (deletion, translation, rotation, treated uniformly as per-frame pose modifications) and physics-aware pose correction driven by road excitation. All Gaussian parameters of dynamic vehicles remain fixed. At render time, primitives in object coordinates are transformed to world coordinates via the updated extrinsic parameters. When the road geometry is edited (Section~\ref{sec:static}), the dynamics module (Section~\ref{sec:dynamic}) samples the road height at ground contact points, solves the vehicle's vertical displacement and pitch angle, and superimposes them onto the ego/target vehicle poses, enabling coupled visual editing and physics-aware pose response.

\subsection{Static Branch: Single-Image Driven Road Geometry Insertion}
\label{sec:static}
Following the static branch of our decoupled framework, we insert typical road structures from a single reference image and text prompt, avoiding the need for 3D assets or retraining. The pipeline comprises three lightweight steps: object texture extraction, monocular depth estimation with local 3D point cloud generation, and scale-corrected Gaussian merging.

\subsubsection{Object Texture Extraction}

To isolate the target road structure from cluttered backgrounds and avoid color bleeding during fusion, we adopt the Grounding DINO \cite{liu2024grounding}+SAM \cite{kirillov2023segment} cascade: DINO is driven by text prompts (such as speed bump/sunken pavement, etc.) to complete target localization, and then SAM obtains fine foreground masks. Based on this, the background is removed and pure object textures are cropped to reduce the impact of edge color mixing on subsequent geometric and visual fusion.

\subsubsection{Depth Estimation and Local 3D Point Cloud Generation}
To lift the 2D object texture to metric 3D geometry from a single view, we use the Depth Anything \cite{lin2025depth} model to perform depth prediction on the texture map and the original scene image synchronously, generating a pixel-wise dense depth map. We use the segmentation mask to remove background and noisy depth values, then normalize and scale the remaining relative depth to absolute metric depth based on preset real-world dimensions.

Given the preset real-world dimensions, 2D pixels are converted into local 3D point-cloud coordinates ($x_{local}$, $y_{local}$, $z_{local}$):
\begin{align}
x_{local}&=\left(\frac{row}{h-1}-0.5\right)L_x \\ 
y_{local}&=\left(\frac{col}{w-1}-0.5\right)L_y \\ 
z_{local}&=\frac{d-d_{min}}{d_{max}-d_{min}}L_z
\label{eq:pointcloud}
\end{align}
where $h$ and $w$ are the height and width of the depth map (identical to those of the cropped object texture), and $row \in [0, h{-}1]$, $col \in [0, w{-}1]$ are the zero-based pixel indices; $L_x$, $L_y$, $L_z$ denote the real-world physical length, width, and height preset for the target road geometry element; $d$ is the per-pixel depth value from Depth Anything within the foreground mask; $d_{min}$ 
and $d_{max}$ are the minimum and maximum depth values across the foreground 
region after background removal.

\subsubsection{Scale Correction and Gaussian Merging for Monocular Depth–derived Point Clouds}
\label{sec:scale_correction}

To resolve the scale mismatch and blurring caused by overly dense monocular depth-derived point clouds, we modify the original 3DGS scale initialization. The inserted geometry is obtained by back-projecting monocular depth to get dense point clouds. The local sampling density far exceeds the sparse SfM point clouds used in the 3DGS training stage. If the original KNN scale initialization formula $log\left(\sqrt{d^2+\epsilon}\right)$ (where $d$ is the Euclidean distance to the nearest neighbor in the point cloud, $\epsilon = 10^{-7}$) is directly applied, dense point clouds will cause excessive overlap of Gaussian primitives in local areas, and obvious color bleeding and blurring artifacts will be generated after $\alpha$-blending during rendering. Therefore, on the premise of retaining neighborhood geometric adaptability, a global scaling factor $\sigma \in (0,1]$ is introduced to uniformly tighten the scale to reduce the effective radiation range and suppress excessive mixing:
\begin{equation}
s=log\left(\sqrt{d^2+\epsilon}\right)+log\left(\sigma\right) \ , \ \sigma=0.01
\label{eq:scale}
\end{equation}

After scale initialization, the remaining Gaussian attributes are assigned as follows and merged into the static background Gaussian set. The three-dimensional point cloud coordinates are rigidly transformed to world coordinates ($X_w, Y_w, Z_w$). Color is represented by spherical harmonics: the zero-order (DC) coefficient is taken from the texture RGB, while higher-order coefficients are set to zero to match the monocular appearance prior and avoid ambiguity under missing multi-view constraints. Opacity is uniformly set to 0.95. The new primitives are merged into the static background Gaussian set and the updated checkpoint is written back, completing road element insertion without retraining.

\subsubsection{Statistical Color Transfer in Lab Space}
To eliminate color discrepancy between the inserted object and the surrounding road surface, we adopt the statistical color transfer method~\cite{reinhard2001color}. The method performs mean-variance matching in Lab space, combined with brightness fine-tuning and linear blending to preserve local texture contrast.

Let the source point cloud RGB be $c_{src}$ and the reference background color be $c_{ref}$ (sampled from the original road surface, clipped to 2\%–98\% quantiles to exclude outliers). After converting both to Lab space, we obtain the per-channel means $(\mu^L,\mu^a,\mu^b)$ and standard deviations $(\sigma^L,\sigma^a,\sigma^b)$. The $a$ and $b$ channels are linearly mapped via:
\begin{align}
a_{new}&=\frac{a_{src}-\mu^a_{src}}{\sigma^a_{src}} \cdot \sigma^a_{ref} + \mu^a_{ref} \\ 
b_{new}&=\frac{b_{src}-\mu^b_{src}}{\sigma^b_{src}} \cdot \sigma^b_{ref} + \mu^b_{ref}
\label{eq:ab_transfer}
\end{align}
To avoid flattening texture contrast, we apply only a mean shift to the $L$ channel:
\begin{equation}
L_{new}=L_{src}+\lambda\left(\mu^L_{ref}-\mu^L_{src}\right)
\label{eq:L_transfer}
\end{equation}
with $\lambda=0.2$. The transferred values are converted back to RGB, and the final color is obtained by linearly blending the source and transferred colors:
\begin{equation}
c_{final}=(1-\beta)c_{src}+\beta \cdot c_{new}
\label{eq:final_color}
\end{equation}
where $\beta=0.75$. This strategy naturally integrates the inserted road element while balancing color consistency and texture retention.
\subsection{Dynamic Branch: Physics-Aware Pose Correction via Vehicle Dynamics}
\label{sec:dynamic}
Besides direct pose editing (deletion, translation, rotation) which simply updates the pose sequences of dynamic vehicles, our framework also supports physics-aware pose correction via vehicle dynamics. Following the dynamic branch of our decoupled framework, edited road geometry provides the excitation input for our vehicle dynamics model. We solve the vehicle body pose in real time, directly correcting the 3DGS rendering pose to couple visual editing with physics-consistent vehicle motion.

\subsubsection{Standard Half-Car 4-DOF Dynamics Model}
\begin{figure}[!tb]
\centering
\includegraphics[width=1.0\linewidth]{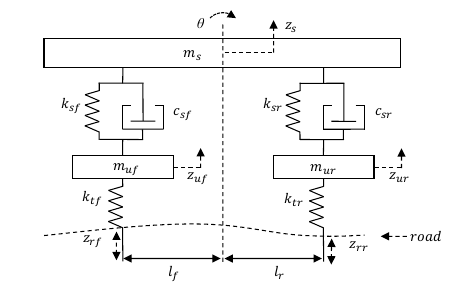}
\caption{Definition of the half-car model.}
\label{fig:halfcar}
\end{figure}

To instantiate the proposed physics-aware correction, we compute dynamics-consistent vertical displacement and pitch angle under road excitation using the classic 4-DOF half-car dynamics model~\cite{rajamani2011vehicle}, illustrated in Fig.~\ref{fig:halfcar}. This model balances accuracy and efficiency, and is widely used in vehicle-road vibration and ride comfort analysis. The model has four independent degrees of freedom: the vertical displacement $z_s$ at the center of mass and the pitch angle $\theta$ of the sprung mass, and the vertical displacements $z_{uf}$, $z_{ur}$ of the front and rear unsprung masses.

Based on Newton's second law and the rigid body rotation law, the system dynamics differential equations are established as follows:
\begin{equation}
\label{eq:vehicle_dynamics}
\left\{
\begin{array}{l}
m_s \ddot{z}_s = -k_{sf}\left(z_s-z_{uf}-l_f\theta\right)-c_{sf}\left(\dot{z}_s-\dot{z}_{uf}-l_f\dot{\theta}\right) \\
\qquad\quad -k_{sr}\left(z_s-z_{ur}+l_r\theta\right)-c_{sr}\left(\dot{z}_s-\dot{z}_{ur}+l_r\dot{\theta}\right) \\[6pt]
I_y \ddot{\theta} = l_f k_{sf}\left(z_s-z_{uf}-l_f\theta\right)+l_f c_{sf}\left(\dot{z}_s-\dot{z}_{uf}-l_f\dot{\theta}\right) \\
\qquad\quad -l_r k_{sr}\left(z_s-z_{ur}+l_r\theta\right)-l_r c_{sr}\left(\dot{z}_s-\dot{z}_{ur}+l_r\dot{\theta}\right) \\[6pt]
m_{uf} \ddot{z}_{uf} = k_{sf}\left(z_s-z_{uf}-l_f\theta\right) + c_{sf}\left(\dot{z}_s-\dot{z}_{uf}-l_f\dot{\theta}\right) \\
\qquad\qquad - k_{tf}\left(z_{uf}-z_{rf}\right) \\[6pt]
m_{ur} \ddot{z}_{ur} = k_{sr}\left(z_s-z_{ur}+l_r\theta\right)+c_{sr}\left(\dot{z}_s -\dot{z}_{ur}+l_r\dot{\theta}\right) \\
\qquad\qquad -k_{tr}\left(z_{ur}-z_{rr}\right)\\[6pt]
\end{array}
\right.
\end{equation}

The symbols and units are defined as follows.
$m_s$ (kg) is the sprung mass,
$I_y$ (kg$\cdot$m$^2$) is the pitch moment of inertia,
$l_f$ and $l_r$ (m) are the distances from the barycenter to the front and rear axles,
$m_{uf}$ and $m_{ur}$ (kg) are the front and rear unsprung masses,
$k_{sf}$ and $k_{sr}$ (N/m) are the front and rear suspension stiffnesses,
$c_{sf}$ and $c_{sr}$ (N$\cdot$s/m) are the front and rear suspension damping coefficients,
and $k_{tf}$ and $k_{tr}$ (N/m) are the front and rear tire stiffnesses.
Dotted and double-dotted variables denote first and second time derivatives.
$z_{rf}$ and $z_{rr}$ (m) are the road heights at the front and rear contact points.

We solve the above differential equations using the fourth-order Runge-Kutta (RK4) method. The resulting vehicle state curves are then sampled at each video frame timestamp and applied as pose corrections to the corresponding Gaussians.

\subsubsection{Road Height Field and Pose Correction}
To couple the edited road geometry with the dynamics model, we build a dense height field from Gaussians and sample contact-point excitations. We first project the edited road Gaussians onto a ground-plane grid and compute a signed height residual with respect to the local road plane. For protruding structures such as speed humps, we retain the maximum residual per cell; for depressions such as sunken roads, we retain the minimum residual per cell.

When the vehicle moves, the heights at the front and rear contact points are sampled frame by frame as the road excitation $z_{rf}$ and $z_{rr}$.
Eq.~\eqref{eq:vehicle_dynamics} is then solved numerically via RK4 to yield the body vertical displacement $z_s$ and pitch angle $\theta$, which are directly superimposed onto the 3DGS vehicle poses to produce physics-aware, dynamics-consistent vertical displacement and pitch responses.

\section{Experiments}
This section evaluates our method through comparative and ablation experiments using both quantitative and qualitative analyses.

\subsection{Experimental Setup}
All experiments are conducted on an NVIDIA RTX 4090 GPU with PyTorch. We evaluate on 10 scenes from the Waymo Open Dataset, inserting two typical road structures (speed humps and sunken roads) for comparison with other editing methods.

\textbf{Evaluation metrics}: 
\begin{itemize}
\item \textbf{CLIP Direction Similarity (CLIPdir)}\cite{xiong2025drivinggaussian++}: Measures the cosine between image feature changes and text prompt feature changes. Higher is better.
\item \textbf{CIEDE2000}\cite{tejada2024exploring}: Quantifies the color difference between edited and reference road surfaces. Lower is better.
\item \textbf{Sharpness}: Laplacian variance and Tenengrad gradient\cite{kong2025progressive} are adopted to quantify edge sharpness. Higher is sharper.
\item \textbf{Dynamic response error}: We report RMSE and extrema error (Ext. Err, peak/trough discrepancy) for pitch angle $\theta$ and vertical displacement $z_s$ against CarSim. Lower is better.
\item \textbf{User study}\cite{chen2024gaussianeditor}: 10 subjects rate visual naturalness on a 1--5 scale (5 = best). Higher is better.
\end{itemize}
\subsection{Comparative Experiments}
We first provide a reference-level comparison with representative 3D editing methods, then evaluate dynamics-response consistency against CarSim and rendering realism via user study.

\subsubsection{Reference-Level Comparison with 3D Editing Methods}

Since existing methods differ in datasets, edit types, hardware, prompt settings, and supported operations, we use Table~\ref{tab:comparison} as a reference-level comparison rather than a controlled benchmark. Adapting appearance-level or object-level editing methods to metric road-height insertion with dynamics would require nontrivial extensions beyond their original scope.

\begin{table*}[!tb]
\centering
\caption{Reference-Level Comparison with Representative 3D Editing Methods}
\begin{tabular}{lcccccc}
\toprule
Method & Application Scenario & Optimization-Free & Physics-Aware Response & Running Time & Devices & CLIPdir$\uparrow$ \\
\midrule
Instruct-NeRF2NeRF \cite{haque2023instruct} & General scenarios & $\times$ & $\times$ & $\approx$\ 60\,min & Titan RTX & 0.16 \\
GaussianEditor \cite{chen2024gaussianeditor} & General scenarios & $\times$ & $\times$ & $\approx$\ 5-10\,min & RTX A6000 & 0.2071 \\
GaussCtrl \cite{wu2024gaussctrl} & General scenarios & $\times$ & $\times$ & $\approx$\ 9\,min & A5000 & 0.192 \\
3DSceneEditor \cite{yan20243dsceneeditor} & General scenarios & $\times$ & $\times$ & $\approx$\ 2-5\,min & A100 & 0.232 \\
GaussianCut \cite{jain2025gaussiancut} & General scenarios & $\checkmark$ & $\times$ & $\approx$\ 1.5\,min & RTX 4090 & -- \\
DrivingGaussian++ \cite{xiong2025drivinggaussian++} & \textbf{Driving scene editing} & $\times$ & $\times$ & $\approx$\ 8\,min & 8$\times$RTX 8000 & 0.2327 \\
Ours & \textbf{Driving scene editing} & $\checkmark$ & $\checkmark$ & $\approx$\ \textbf{6.24\,s} & RTX 4090 & \textbf{0.257} \\
\bottomrule
\multicolumn{7}{p{\dimexpr\textwidth-2\tabcolsep\relax}}{\footnotesize Note: All comparison values (CLIPdir, Running Time, Devices) are taken or derived from the respective original papers. Since hardware varies across methods, this table serves as a reference-level comparison. The GaussCtrl CLIPdir value is averaged over multiple scenes as reported in \cite{wu2024gaussctrl}.}
\end{tabular}
\label{tab:comparison}
\end{table*}

Gaussian Grouping \cite{ye2024gaussian}, HorizonForge \cite{wang2026horizonforge}, and HorizonWeaver \cite{liu2026horizonweaver} are omitted due to unavailable comparable metrics or different edit settings. RoVES instead targets physics-aware road-height insertion coupled with vehicle dynamics for vertical displacement and pitch responses.

\textbf{Efficiency analysis}: Table~\ref{tab:runtime} shows that object insertion takes 1.84\,s on our platform, and the full pipeline takes 6.24\,s. Although cross-paper runtimes in Table~\ref{tab:comparison} are not directly comparable, our breakdown shows that RoVES completes road editing and vehicle-dynamics-based pose correction within seconds, supporting rapid batch generation.

\begin{table}[!tb]
\centering
\caption{Runtime Breakdown of the Proposed Editing Pipeline}
\setlength{\tabcolsep}{5pt}
\begin{tabular}{lc}
\toprule
Module & Running Time \\
\midrule
Texture extraction & 0.04\,s \\
Depth estimation & 0.94\,s \\
Point cloud generation & 0.09\,s \\
Statistical Lab color transfer (optional) & 2.67\,s \\
Gaussian primitive initialization and scene merging & 0.77\,s \\
Vehicle-dynamics solving and pose correction & 1.73\,s \\
\midrule
Total & 6.24\,s \\
\bottomrule
\end{tabular}
\label{tab:runtime}
\end{table}

\subsubsection{Evaluation of Dynamic Response and Rendering Realism}
We evaluate two key aspects of our framework: the response consistency of the half-car dynamics model against CarSim, and the visual realism of the rendered scenes with vehicle-dynamics-based pose correction (evaluated via user study). For dynamics validation, we select Scene 150 from the 10 edited Waymo scenes as a representative case and insert a sunken road. We then simulate a front vehicle (light truck) followed by the ego vehicle (sedan) traversing the edited road. Vehicle parameters are listed in Table~\ref{tab:vehicle_params}, and the same road profile and vehicle parameters are used for both RoVES and CarSim.

\begin{table}[!tb]
\centering
\caption{Vehicle Parameters Used in Dynamics Simulation}
\setlength{\tabcolsep}{6pt}
\begin{tabular}{lccc}
\toprule
Parameters      & Ego Vehicle & Front Vehicle & Unit  \\
\midrule
$m_s$           & 1200        & 2600           & kg    \\
$I_y$           & 1800        & 4800           & kg$\cdot$m² \\
$l_f$           & 1.2         & 1.1            & m     \\
$l_r$           & 1.5         & 1.9            & m     \\
$m_{uf}/m_{ur}$ & 54          & 110            & kg    \\
$k_{sf}/k_{sr}$ & 18000       & 52000          & N/m   \\
$c_{sf}/c_{sr}$ & 3200        & 6500           & N$\cdot$s/m \\
$k_{tf}/k_{tr}$ & 180000      & 380000         & N/m   \\
\bottomrule
\multicolumn{4}{l}{Note: Derived from typical ranges in the literature~\cite{rajamani2011vehicle}.}
\end{tabular}
\label{tab:vehicle_params}
\end{table}

\begin{figure}[htbp]
\centering
\includegraphics[width=1\columnwidth]{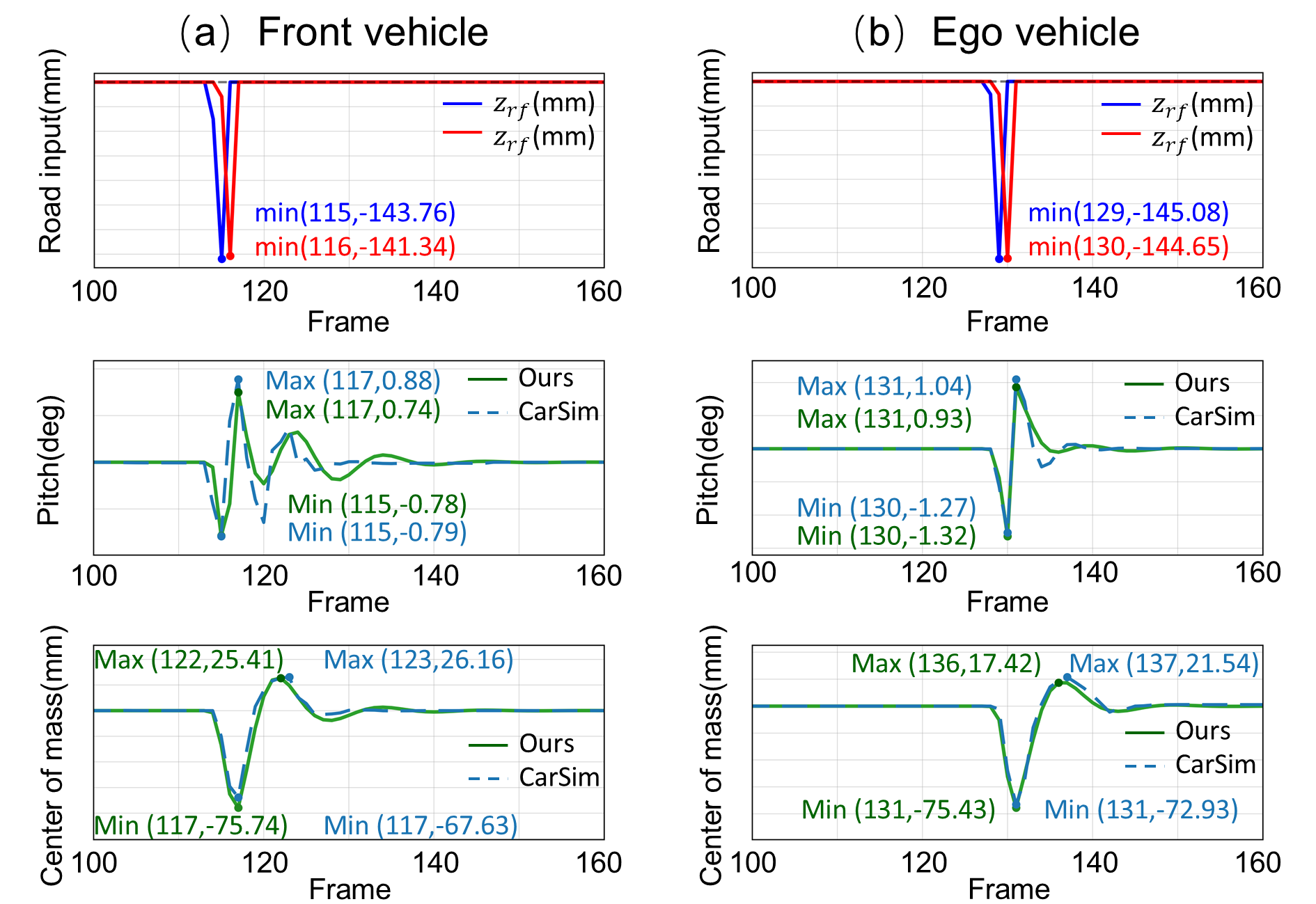}
\caption{Comparison of pitch and vertical response trends between our model and CarSim. The ``No dynamics'' case (zero curves) is omitted.}
\label{fig:dynamic_response}
\end{figure}

\textbf{Quantitative comparison.} Fig.~\ref{fig:dynamic_response} compares the time-domain responses of our model (ego and front vehicles) against the CarSim reference, and Table~\ref{tab:carsim_metrics} reports the corresponding error metrics. For pitch angle, the extrema error (Ext. Err.) is below 0.14$^\circ$, with RMSEs of 0.125$^\circ$ and 0.055$^\circ$ for the front and ego vehicles, respectively. For the vertical displacement of the center of mass, the extrema error is below 8.2\,mm, with RMSEs of 2.87\,mm and 1.87\,mm, respectively. These results indicate that our half-car solver closely matches the CarSim response in both amplitude and temporal trend.

\begin{figure}[!tb]
\centering
\includegraphics[width=1 \linewidth]{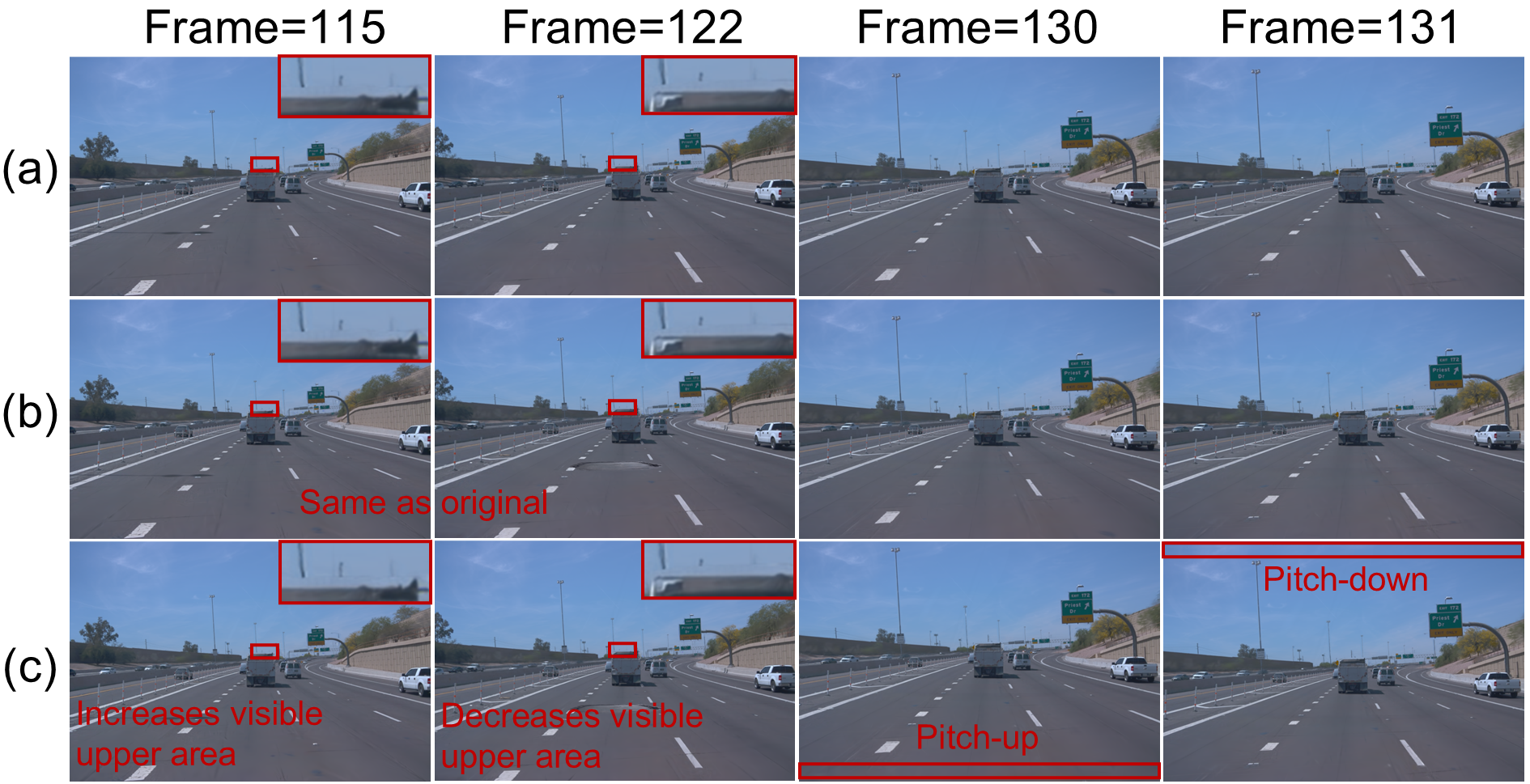}
\caption{Visual comparison of rendered driving scenes. (a) Original scene; (b) Rendering without vehicle-dynamics-based pose correction; (c) Rendering with physics-aware pose correction (Ours).}
\label{fig:visual_comparison}
\end{figure}

\textbf{Qualitative comparison (rendering realism).} Fig.~\ref{fig:visual_comparison} shows the rendered driving scenes with and without vehicle-dynamics-based pose correction. The physics-aware correction produces visually plausible pitch and vertical displacement (user study score 4.7), while the uncorrected case lacks plausible vehicle response (user study 2.1), confirming the necessity of our dynamics module.

\begin{table}[!tb]
\centering
\caption{Dynamic Response Agreement with CarSim}
\setlength{\tabcolsep}{3pt}
\begin{tabular}{lcccc}
\toprule
Vehicle & Pitch RMSE & Pitch Ext. Err. & $z_s$ RMSE & $z_s$ Ext. Err. \\
        & (deg)      & (deg)           & (mm)       & (mm) \\
\midrule
Front vehicle & 0.125 & 0.139 & 2.87 & 8.11 \\
Ego vehicle   & 0.055 & 0.114 & 1.87 & 4.13 \\
\bottomrule
\end{tabular}
\label{tab:carsim_metrics}
\end{table}

\subsection{Ablation Experiments}
We further ablate the KNN adaptive scale factor $\sigma$ and the color transfer module, as these components directly affect rendering sharpness and visual consistency.

\subsubsection{Ablation Experiment on KNN Adaptive Scale Control Scaling Factor}
We evaluate the impact of the global scaling factor $\sigma$ (see Sec.~\ref{sec:scale_correction}) on the rendering sharpness of inserted objects. Speed humps of various textures are inserted into the same scene, and we test $\boldsymbol{\sigma \in \{1.0, 0.5, 0.1, 0.01\}}$, where $\sigma=1.0$ reproduces the original 3DGS scale initialization as a fair baseline.

\begin{table}[t]
\centering
\caption{Ablation Study on Gaussian Scale Parameter $\sigma$}
\setlength{\tabcolsep}{3pt}
\begin{tabular}{lccc}
\toprule
$\sigma$ & Laplacian Variance$\uparrow$ & Tenengrad Gradient$\uparrow$ & User Study$\uparrow$ \\
\midrule
1 (Original) & 113.14 & 9823.82 & 3.8 \\
0.5 & 376.99 & 15856.15 & 4.2 \\
0.1 & 666.07 & 18313.65 & \textbf{4.5} \\
0.01 (Ours) & \textbf{700.35} & \textbf{18483.72} & \textbf{4.5} \\
\bottomrule
\multicolumn{4}{p{0.95\columnwidth}}{Note: All values are averaged over three speed hump instances. $\sigma=0.001$ yields results highly similar to $\sigma=0.01$ (both quantitatively and qualitatively) and is therefore omitted.}
\end{tabular}%
\label{tab:sigma_ablation}
\end{table}

\begin{figure}[!tb]
\centering
\includegraphics[width=1.0\linewidth]{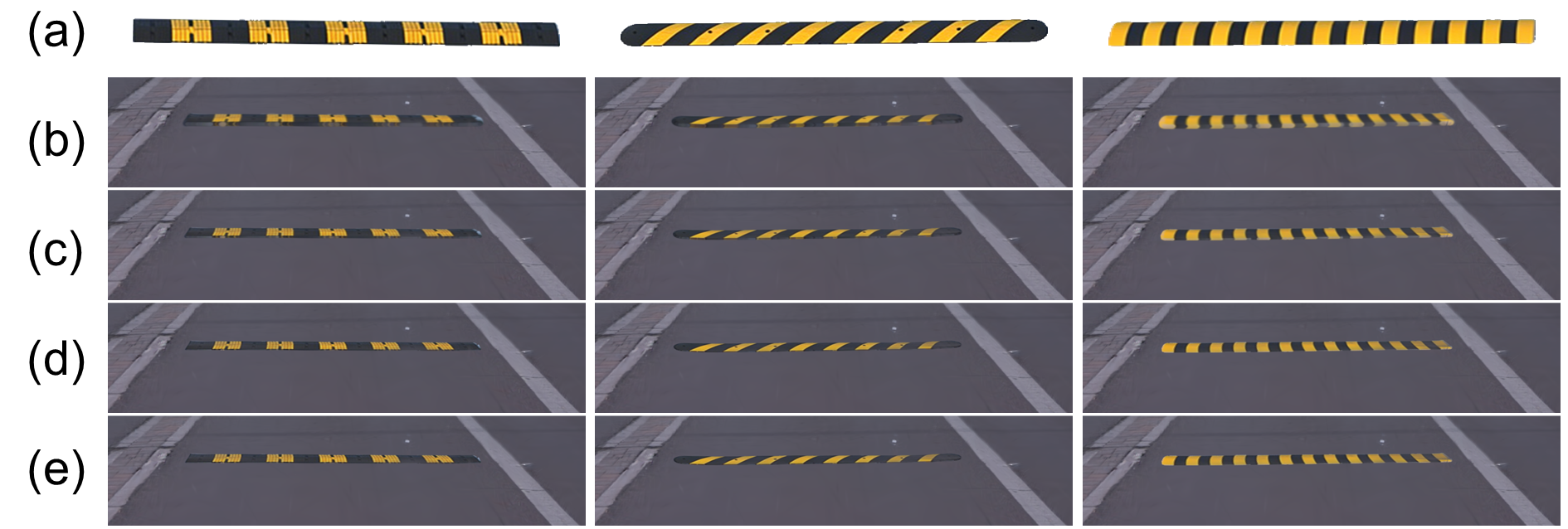}
\caption{Visualization of the effect of Gaussian scale parameter $\sigma$ on rendering sharpness. (a) Input textures; (b)-(e) Rendering results with $\sigma = 1, 0.5, 0.1, 0.01$ (from top to bottom), respectively.}
\label{fig:sigma_visual}
\end{figure}

\textbf{Quantitative analysis.} Table~\ref{tab:sigma_ablation} reports the Laplacian variance and Tenengrad gradient under different scaling factors. Our KNN adaptive strategy ($\sigma=0.01$) achieves the highest scores on both metrics, significantly outperforming the original 3DGS initialization ($\sigma=1.0$) across all three speed hump instances. This demonstrates effective alleviation of the scale mismatch caused by dense monocular point cloud insertion.

\textbf{Qualitative analysis.} Fig.~\ref{fig:sigma_visual} shows the local magnification comparison. The original 3DGS ($\sigma=1.0$) produces blurred edges and lost texture details; $\sigma=0.5$ improves edge sharpness with slight residual blurring; $\sigma=0.01$ and $\sigma=0.1$ both yield clear edges and distinct textures with negligible visual difference. Considering the numerical improvement and parameter robustness, we select $\sigma=0.01$ as the default value.

\subsubsection{Ablation Experiment on Color Transfer}
To verify the effectiveness of the color transfer module on visual consistency after sunken road insertion, we compare four strategies that vary by \textbf{reference source (point cloud vs.\ rendered image) and color space (RGB vs.\ Lab)}. The quantitative results are reported in Table~\ref{tab:color_ablation}.

\begin{table}[!tb]
\centering
\caption{Quantitative Comparison of Different Color Transfer Strategies}
\label{tab:color_ablation}
\begin{tabular}{lcccc}
\toprule
Method & PC-RGB & PC-Lab & Img-RGB & Img-Lab (Ours) \\
\midrule
CLIPdir$\uparrow$ & 0.2256 & 0.2175 & 0.0564 & \textbf{0.2631} \\
CIEDE2000$\downarrow$ & 17.9592 & 6.6190 & \textbf{1.7976} & 3.0623 \\
User Study $\uparrow$ & 2.52 & 3.82 & 2.73 & \textbf{4.16} \\
\bottomrule
\multicolumn{5}{l}{Note: PC/Img: reference source; RGB/Lab: color space for transfer.}
\end{tabular}
\end{table}

\begin{figure}[!tb]
\centering
\includegraphics[width=1\linewidth]{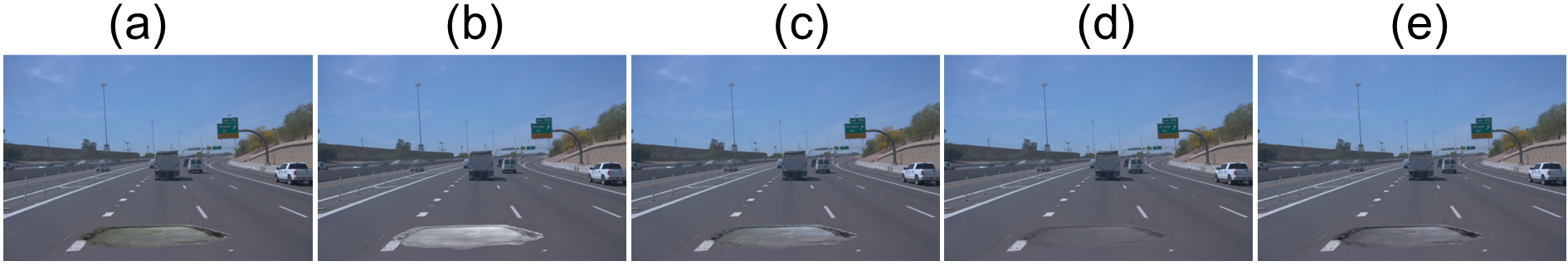}
\caption{Color transfer results under different strategies. (a) No color transfer; (b) PC-RGB; (c) PC-Lab; (d) Img-RGB; (e) Img-Lab (Ours).}
\label{fig:color_visual}
\end{figure}

\textbf{Quantitative analysis.} As shown in Table~\ref{tab:color_ablation}, PC-RGB performs worst overall, while PC-Lab achieves higher user ratings but with larger color difference. Img-RGB yields the lowest CIEDE2000 at the cost of semantic alignment. Img-Lab (Ours) obtains the best balance across all three metrics, validating its advantage in perceptual consistency and texture preservation.

\textbf{Qualitative analysis.} Fig.~\ref{fig:color_visual} shows the color transfer results under different strategies. PC-RGB produces obvious color discrepancy from the background road; PC-Lab yields visually acceptable results but retains noticeable color differences; Img-RGB suppresses color discrepancy at the cost of texture detail; Img-Lab (Ours) naturally blends the inserted object with the road while preserving local contrast and texture.

\section{Conclusion}

We presented RoVES, a novel physics-aware 3D Gaussian editing framework for driving scene generation. By decoupling static Gaussian insertion from dynamic pose updates and coupling edited road geometry with a half-car vehicle dynamics model, RoVES unifies visual scene editing with physics-aware vehicle pose correction. It achieves second-level editing (6.24\,s) while producing dynamics-consistent vertical displacement and pitch responses. Experiments on the Waymo dataset suggest that RoVES provides a practical step toward bridging 3DGS-based visual scene editing and physics-aware driving scene generation, particularly for road geometry editing, vehicle-response modeling, and efficient batch generation. 

\section{Limitations and Future Work}

The current physics-aware module considers only vertical vehicle dynamics, including vertical displacement and pitch, and requires manually specified vehicle parameters. This limits its direct applicability to diverse vehicle types without per-vehicle tuning. Moreover, when edited road structures induce extreme vehicle pose deviations, the resulting viewpoints can fall outside the well-reconstructed regions of the original 3DGS scene, leading to rendering artifacts that degrade visual quality. Future work will address these limitations by automating vehicle parameter identification from visual cues and extending the model with lateral dynamics for more complete vehicle motion modeling, and by exploring diffusion-based inpainting or generation to faithfully restore artifact regions under large-amplitude viewpoint changes, while preserving the one-shot nature of 3DGS insertion.

\bibliographystyle{IEEEtran}
\balance

\end{document}